\documentclass{article}

\usepackage[dblblindworkshop,final]{neurips_2026}
\workshoptitle{MusIML@NeurIPS 2026}

\usepackage[utf8]{inputenc}
\usepackage[T1]{fontenc}
\usepackage{hyperref}
\hypersetup{
  hypertexnames=false,
  hidelinks
}
\usepackage{url}
\usepackage{booktabs}
\usepackage{graphicx}
\usepackage[table]{xcolor}
\usepackage{multirow}
\usepackage{enumitem}
\usepackage{float}
\usepackage{fancyvrb}
\definecolor{tablegray}{gray}{0.90}
\definecolor{stripegray}{gray}{0.94}
\definecolor{rulegray}{gray}{0.35}
\definecolor{closedgray}{gray}{0.40}
\definecolor{openteal}{RGB}{42,157,143}
\definecolor{linkblue}{RGB}{0,64,160}

\newcommand{\githublink}[2]{\href{#1}{\textcolor{linkblue}{\textbf{GitHub/#2}}}}

\setlist[itemize]{topsep=2pt,itemsep=1pt,parsep=0pt,leftmargin=*}
\newcommand{\safecitep}[1]{\citep{#1}}

\makeatletter
\renewcommand{\section}{%
  \@startsection{section}{1}{\z@}%
                {-1.3ex}%
                {0.55ex}%
                {\large\bf\raggedright}%
}
\renewcommand{\paragraph}{%
  \@startsection{paragraph}{4}{\z@}%
                {0.45ex}%
                {-1em}%
                {\normalsize\bf}%
}
\makeatother

\title{When Models Misread Ink: Auditing VLMs for Arabic Calligraphy Recognition}

\author{
Niaz Mohaiman Abtahi\textsuperscript{1},
Evan Maksud\textsuperscript{1},
Reyana Shahid\textsuperscript{2}\\
\textsuperscript{1}Advanced AI Lab Limited\\
\textsuperscript{2}United International University\\
Correspondence to: \texttt{tahiniaz@gmail.com}\\[0.35em]
\githublink{https://github.com/NiazTahi/When-Models-Misread-Ink}{When-Models-Misread-Ink}\\[-1.0em]
}

\begin{document}

\maketitle

\begin{abstract}
Arabic calligraphy is a high-stakes cultural visual domain for Muslim communities, yet VLM evaluations rarely test its fine-grained script traditions.
We audit 12 VLMs on a cleaned 1,195-image six-way benchmark using numeric, visual-cue, and Diwani-first prompts.
In this closed-set setting, the best numeric-prompt accuracy is 60.9\% for closed/API models and 47.2\% for open-weight models.
The main failure is default-label collapse: every numeric run is dominated by a single label taking 33--63\% of predictions, against an 18.8\% majority-class baseline.
For 8 of 12 models the default is Thuluth; GPT-4o labels 52.9\% of all images as Thuluth although it is only 18.1\% of the set.
Prompting helps unevenly: visual cues move accuracy by more than 10 points for some models but hurt others, and foregrounding Diwani often relocates the default rather than removing it.
Arabic calligraphy recognition is therefore a compact, culturally grounded stress test for VLM robustness and for measuring which visual traditions become legible to automated systems.
\end{abstract}

\section{Introduction}

This paper studies a narrow but important question: given an image of Arabic calligraphy, can modern vision--language models identify its script style rather than collapse it into a familiar label?
More specifically, when the choices are Kufic, Naskh, Thuluth, Diwani, Ruq'ah, and Nasta'liq, do models show stable visual discrimination, or do prompt wording and label order change which traditions become legible?

This question sits at the intersection of rapid VLM progress~\safecitep{radford2021clip,alayrac2022flamingo,liu2023llava}, culturally diverse vision evaluation~\safecitep{liu2021marvl,romero2024cvqa,ramaswamy2023geode}, and evidence that model outputs can shift with option order and prompt format~\safecitep{zheng2024mcqbias,pezeshkpour2024order,sclar2024formatspread,lu2022promptorder}.
Prior work introduced DuwatBench for Arabic calligraphy understanding~\safecitep{patle2026duwatbench}; we focus on whether VLMs have reliable style priors after rebalancing the evaluation set and testing controlled prompt interventions.
The task is culturally grounded for Muslim communities~\safecitep{musiml2026}: calligraphic styles carry historical, religious, regional, and institutional associations, so systematic misclassification can distort access to visual heritage.

Our contributions are:
\begin{itemize}
    \item a cleaned and rebalanced 1,195-image benchmark for six-way Arabic calligraphy style recognition;
    \item a controlled evaluation of four closed/API VLMs and eight open-weight VLMs on the same image manifest;
    \item three prompt conditions---numeric, visual-cue, and Diwani-first---that test whether style descriptions or explicit attention to an under-recognized class repair errors; and
    \item analyses of default-label collapse, Thuluth over-prediction, Diwani under-recognition, and paired prompt effects.
\end{itemize}

Our central finding is not simply that current VLMs are inaccurate.
Rather, their errors are culturally structured: many models overuse Thuluth as a default, fail to recover Diwani, and respond unevenly to prompts that should make the visual distinctions clearer.

\section{Dataset and Task}

\paragraph{Dataset.}
We use a cleaned 1,195-image six-style dataset.
DuwatBench~\safecitep{patle2026duwatbench} is supplemented with RuFa images for Ruq'ah/Nasta'liq~\safecitep{rufa} and HICMA images for Kufic/Naskh~\safecitep{hicma}, reducing the original Thuluth-heavy imbalance.
Curation uses source aggregation, single-label filtering, class-balance supplementation, file-integrity checks, duplicate removal, and manifest locking.
After removing one undersized image and four exact duplicates, the final counts are Kufic 106, Naskh 218, Thuluth 216, Diwani 217, Ruq'ah 225, and Nasta'liq 213; the majority baseline falls from 56.4\% to 18.8\%.
Thuluth and Diwani are both entirely DuwatBench, so their central confusion is not cross-source domain shift; other labels use mixed sources, disclosed in the appendix.

\paragraph{Task.}
The task is closed-set six-way style classification: given one image, select one script style.
We evaluate style recognition only, not OCR, transcription, semantic interpretation, religious-content judgment, or localization.
All metrics use the same 1,195-image manifest for every model--prompt run.

\section{Experimental Setup}

\paragraph{Prompt conditions.}
We use three prompts.
\textbf{Numeric} lists numbered styles and requires one digit.
\textbf{Visual-cue} adds short style descriptions.
\textbf{Diwani-first} keeps the visual-cue framing but moves Diwani first and asks for the script name, testing whether foregrounding an under-recognized style rescues Diwani errors.
We introduced this condition after early runs showed Diwani was often missed or absorbed into Thuluth; it is a diagnostic prompt, not a preferred deployment prompt.

\paragraph{Models.}
We evaluate four closed/API models~\safecitep{openai2026models}: GPT-4o, GPT-4.1, GPT-5.5, and GPT-5.6 Terra.
The open-weight set includes Qwen2.5-VL-7B/32B~\safecitep{bai2025qwen25vl}, Qwen3-VL-8B/32B~\safecitep{bai2025qwen3vl}, InternVL3.5-8B/38B~\safecitep{wang2025internvl35}, MiniCPM-V~4.5~\safecitep{yu2025minicpmv45}, and Gemma~4~12B-IT~\safecitep{gemmateam2026gemma4}.
Open-weight runs use greedy decoding with seed 42; closed/API runs use the most deterministic supported settings.

\paragraph{Metrics.}
We report accuracy, balanced accuracy, macro-F1, per-style precision/recall/F1, and prediction ratio: predicted count divided by true count.
The main diagnostics are Thuluth prediction ratio and Diwani recall, because early runs absorbed ornate or uncertain examples into Thuluth while missing Diwani.
Paired prompt comparisons use exact McNemar tests on same-image correctness vectors; visual-cue changes with $p \ge 0.05$ are marked not significant.

\section{Results}

Table~\ref{tab:numeric} summarizes the numeric-prompt comparison.
The best closed/API numeric-prompt accuracy is 60.9\% (GPT-5.5), while the best open-weight numeric-prompt accuracy is 47.2\% (Gemma~4~12B-IT).
Thus, although the task is visually compact and closed-set, it remains difficult for current VLMs.

\begin{table}[H]
\centering
\caption{Numeric-prompt results. Thuluth bias is predicted/true Thuluth count; closer to 1.0 is better. Bold marks the best value within each model block.}
\label{tab:numeric}
\scriptsize
\setlength{\tabcolsep}{3.0pt}
\renewcommand{\arraystretch}{0.88}
\arrayrulecolor{rulegray}
\begin{tabular}{@{}lrrrrrr@{}}
\toprule
Model & Acc. $\uparrow$ & Macro F1 $\uparrow$ & \shortstack{Thuluth\\bias} & \shortstack{Diwani\\recall $\uparrow$} & \shortstack{Diwani\\precision $\uparrow$} & \shortstack{Diwani\\F1 $\uparrow$} \\
\midrule
\rowcolor{tablegray}
\multicolumn{7}{@{}l}{\textit{Closed/API VLMs}} \\
GPT-4o & 50.5 & 0.504 & 2.93 & 15.7 & 75.6 & 0.260 \\
\rowcolor{stripegray}
GPT-4.1 & 55.3 & 0.507 & \textbf{2.05} & 1.8 & 66.7 & 0.036 \\
GPT-5.5 & \textbf{60.9} & 0.616 & \textbf{2.05} & 31.8 & \textbf{71.9} & 0.441 \\
\rowcolor{stripegray}
GPT-5.6 Terra & 60.3 & \textbf{0.628} & 2.07 & \textbf{46.5} & 66.9 & \textbf{0.549} \\
\midrule
\rowcolor{tablegray}
\multicolumn{7}{@{}l}{\textit{Open-weight VLMs}} \\
Qwen2.5-VL-7B & 29.2 & 0.255 & \textbf{0.91} & 9.2 & 26.3 & 0.137 \\
\rowcolor{stripegray}
Qwen2.5-VL-32B & 36.7 & 0.378 & 2.75 & 35.0 & 25.6 & 0.296 \\
Qwen3-VL-8B & 46.5 & \textbf{0.457} & 2.69 & 18.9 & \textbf{63.1} & 0.291 \\
\rowcolor{stripegray}
Qwen3-VL-32B & 40.8 & 0.421 & 3.47 & 6.5 & 53.8 & 0.115 \\
InternVL3.5-8B & 22.9 & 0.246 & 1.33 & 5.5 & 85.7 & 0.104 \\
\rowcolor{stripegray}
InternVL3.5-38B & 35.8 & 0.370 & 1.53 & \textbf{56.2} & 30.5 & \textbf{0.395} \\
MiniCPM-V 4.5 & 31.6 & 0.331 & 1.17 & 42.4 & 18.6 & 0.258 \\
\rowcolor{stripegray}
Gemma 4 12B-IT & \textbf{47.2} & 0.454 & 2.89 & 6.5 & 46.7 & 0.113 \\
\bottomrule
\end{tabular}
\arrayrulecolor{black}
\end{table}

\paragraph{Structured errors.}
The dominant failure is default-label collapse, not random confusion.
Thuluth is over-predicted in 11 of 12 numeric-prompt runs; closed/API models choose it 2.05--2.93$\times$ its true frequency, and Qwen3-VL-32B reaches 3.47$\times$.
This persists after rebalancing: every numeric run has one label taking at least one-third of predictions, rising to 52.9\% for GPT-4o and 62.8\% for Qwen3-VL-32B.
False Thuluth predictions most often come from true Diwani (40.9\%), then Naskh (21.5\%) and Nasta'liq (16.4\%), suggesting both visual-neighbor confusion and an uncertainty default.

\begin{figure}[H]
\centering
\includegraphics[width=0.86\linewidth]{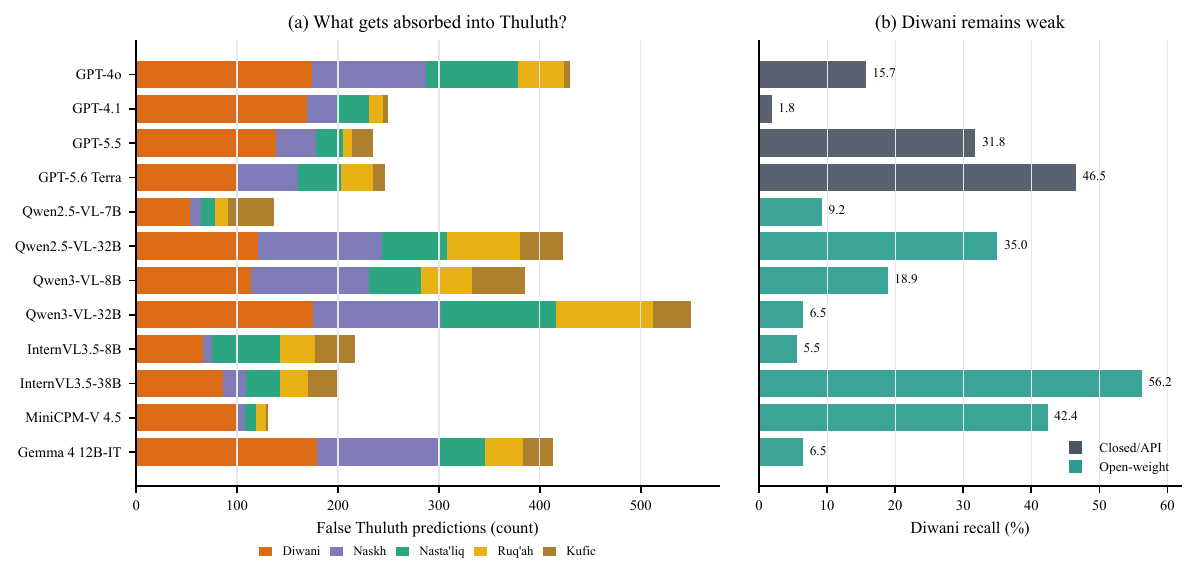}
\caption{Numeric-prompt failure structure. Left: stacked counts show which true labels are incorrectly absorbed into Thuluth. Right: Diwani recall remains low for many models despite higher overall accuracy.}
\label{fig:thuluth_diwani}
\end{figure}

Diwani remains the clearest casualty.
GPT-4.1 recalls only 1.8\% of Diwani images, while Qwen3-VL-32B and Gemma~4~12B-IT recall only 6.5\%; InternVL3.5-38B reaches 56.2\% recall but only 30.5\% precision.
GPT-5.6 Terra is the only model with both non-trivial recall and stronger Diwani F1 (0.549).
Figure~\ref{fig:thuluth_diwani} visualizes this pair of errors: which true labels are absorbed into Thuluth and how much true Diwani each model recovers.

\begin{figure}[H]
\centering
\includegraphics[width=0.83\linewidth]{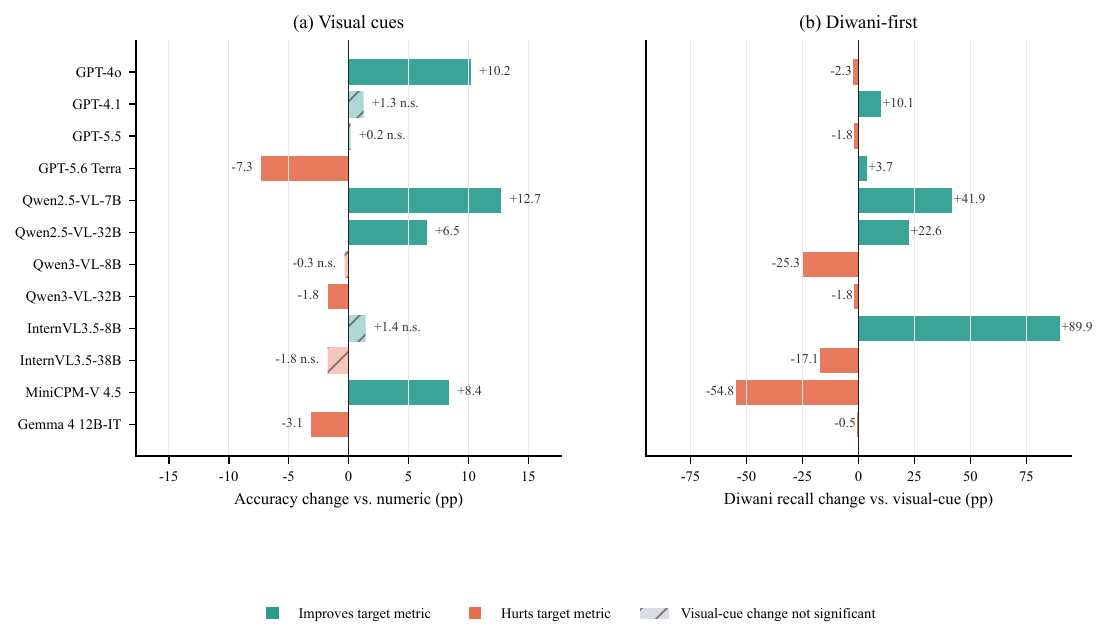}
\caption{Prompt effects relative to the numeric prompt. Visual cues are inconsistent; Diwani-first often raises Diwani recall at the cost of calibration.}
\label{fig:prompt_effects}
\end{figure}

\paragraph{Prompt interventions are unstable.}
Figure~\ref{fig:prompt_effects} shows that prompting does not behave like a uniform fix.
Visual cues improve some models, e.g., GPT-4o (+10.2 points) and Qwen2.5-VL-7B (+12.7), but hurt others, e.g., GPT-5.6 Terra (-7.3) and Gemma~4~12B-IT (-3.1); five of twelve changes are not significant.
Diwani-first can raise recall sharply, especially for InternVL3.5-8B (+89.9), but often relocates the default rather than repairing recognition: InternVL3.5-8B answers Diwani for 90.3\% of all images, while Qwen3-VL models and Gemma~4~12B-IT still end with 0\% Diwani recall.

\section{Discussion and Limitations}

This study supports a concrete claim: Arabic calligraphy recognition remains a brittle fine-grained cultural vision task for current VLMs.
Aggregate accuracy hides the main failure mode: models often behave like default-label emitters with partial discrimination layered on top.
Several models overuse Thuluth as the default label, while others collapse onto Naskh, Diwani, or Nasta'liq depending on model family and prompt.
These errors matter because calligraphic styles encode cultural, historical, and religious contexts, and automated tools should not collapse less-visible traditions into more familiar labels.

\paragraph{Community relevance.}
The failure mode matters for education, museums, manuscript archives, social media, and community-facing search: collapsing Diwani, Ruq'ah, or Nasta'liq into Thuluth can distort access to Islamic visual heritage.

The study has limitations: it covers six broad labels rather than regional schools, mixed-style compositions, dating, handwriting, OCR, or localization; source and label are partly confounded for non-DuwatBench supplements; Prompt C confounds label order and response format; and API-served models can change over time.

Methodologically, the benchmark is useful because it separates two effects that are often blurred in cultural-vision evaluations.
The rebalanced manifest reduces the chance that high Thuluth prediction is simply learned from label frequency, while the prompt interventions show whether the same images become more legible when models are given style descriptions or an explicit Diwani prior.
The negative result is therefore informative: even when a prompt names the right visual features, many systems still substitute a familiar-looking script for the correct one.

\paragraph{Implications.}
For deployment, prompt descriptions are not enough: systems that classify or retrieve Arabic calligraphy should report uncertainty, validate performance per style, and avoid treating high aggregate accuracy as evidence that all traditions are equally legible.

\section{Conclusion}

We present a 12-model evaluation of Arabic calligraphy style recognition under three prompt conditions.
The results reveal limited accuracy, recurring Thuluth over-prediction, Diwani under-recognition, and nonuniform prompt sensitivity across closed/API and open-weight VLMs.
Arabic calligraphy therefore offers a compact culturally grounded benchmark for measuring not only model accuracy, but how errors redistribute visibility across significant visual traditions.

\paragraph{Acknowledgments.}
This work received no external funding. The authors declare no competing interests.

\clearpage
\begingroup
\footnotesize
\setlength{\bibsep}{1pt}
\bibliographystyle{plainnat}
\bibliography{references}
\endgroup

\clearpage
\appendix
\renewcommand{\theHtable}{\Alph{section}.\arabic{table}}
\renewcommand{\theHfigure}{\Alph{section}.\arabic{figure}}

\section{Dataset construction details}

\begin{figure}[H]
\centering
\includegraphics[width=0.86\linewidth]{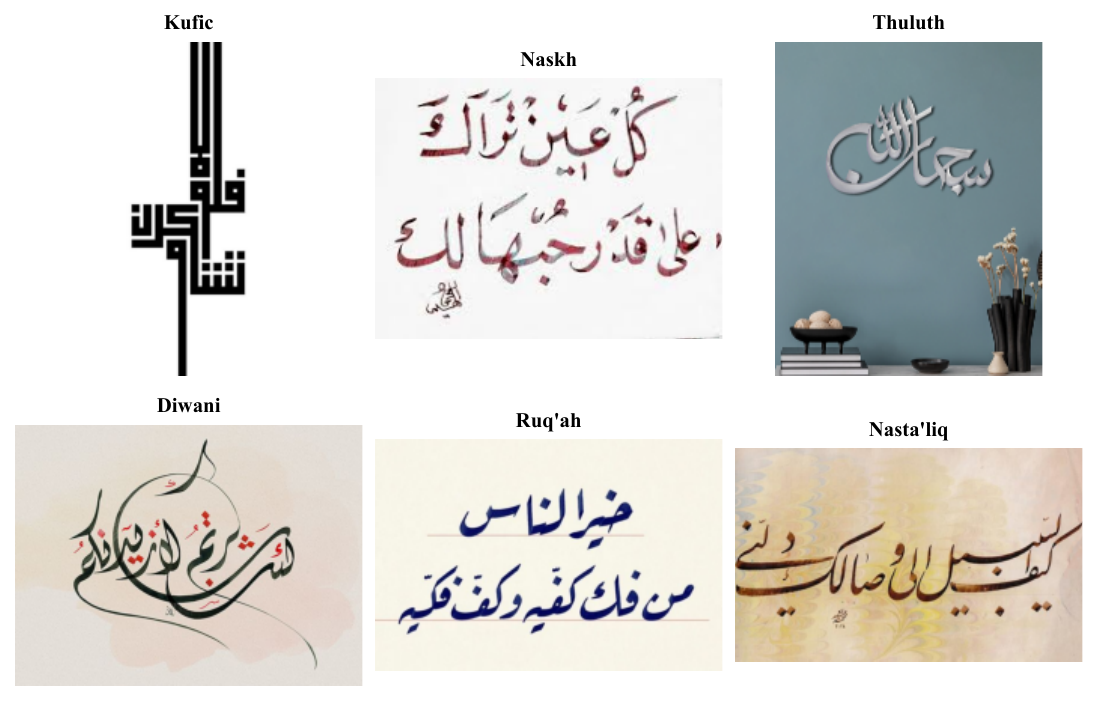}
\caption{Representative examples from the six calligraphy styles in the cleaned evaluation set.}
\label{fig:dataset_examples}
\end{figure}

The evaluated dataset contains 1,195 single-label Arabic calligraphy images.
It was constructed from DuwatBench~\safecitep{patle2026duwatbench} plus real-image supplements from RuFa~\safecitep{rufa} and HICMA~\safecitep{hicma} used to reduce the original class imbalance.
Curation followed a fixed pipeline: aggregate source images, retain single-label six-style items, supplement underrepresented classes, check image-file integrity, remove exact duplicates and undersized files, then lock the manifest used for all inference runs.
The final manifest contains 106 Kufic, 218 Naskh, 216 Thuluth, 217 Diwani, 225 Ruq'ah, and 213 Nasta'liq images.
Table~\ref{tab:source_class} reports the source composition by label.
Thuluth and Diwani share the same DuwatBench provenance, so their central confusion in the main results is not a cross-source artifact; other labels should be interpreted with the mixed-source composition in mind.
Before model evaluation, the dataset was checked for missing files, corrupted images, orphan images, undersized images, and exact duplicates.
One undersized image and four duplicate images were removed.
Figure~\ref{fig:dataset_examples} shows representative examples from the final six-style evaluation set.

\begin{table}[H]
\centering
\caption{Source composition of the cleaned evaluation set by true label.}
\label{tab:source_class}
\scriptsize
\setlength{\tabcolsep}{5pt}
\renewcommand{\arraystretch}{0.95}
\rowcolors{2}{tablegray}{white}
\begin{tabular}{@{}lrrr@{}}
\toprule
True label & DuwatBench & RuFa & HICMA \\
\midrule
Kufic & 80 & 0 & 26 \\
Naskh & 103 & 0 & 115 \\
Thuluth & 216 & 0 & 0 \\
Diwani & 217 & 0 & 0 \\
Ruq'ah & 75 & 150 & 0 \\
Nasta'liq & 64 & 149 & 0 \\
\bottomrule
\end{tabular}
\end{table}

\section{Prompts}

\noindent The prompt text below is line-wrapped for page layout.

\paragraph{Numeric prompt.}
\begin{Verbatim}[fontsize=\scriptsize,baselinestretch=0.95]
You are shown an Arabic calligraphy image. Below are six possible styles,
each with a number:

1. Kufic
2. Naskh
3. Thuluth
4. Diwani
5. Ruq'ah
6. Nasta'liq

Respond with only the number of the style that best matches this image.
Respond with a single digit only -- no words, no explanation.
\end{Verbatim}

\paragraph{Visual-cue prompt.}
\begin{Verbatim}[fontsize=\tiny,baselinestretch=0.95]
You are shown an Arabic calligraphy image. Classify its calligraphic style.

Choose exactly one of these six styles:

1. Kufic
2. Naskh
3. Thuluth
4. Diwani
5. Ruq'ah
6. Nasta'liq

Use the following visual cues:

- Kufic: angular, geometric, squared or block-like letterforms; strong
  horizontal/vertical structure; often rigid and monumental.
- Naskh: clear, balanced, readable manuscript style; rounded letters;
  moderate stroke contrast; commonly used for Qur'anic and book text.
- Thuluth: large, elegant, sweeping curves; elongated verticals; complex
  decorative composition; often used in architectural or ceremonial calligraphy.
- Diwani: highly cursive and ornamental; dense flowing curves; letters may
  interlace; often compact and decorative.
- Ruq'ah: simple, fast handwriting-like style; compact letters; short strokes;
  practical and less ornamental.
- Nasta'liq: diagonal/slanted baseline; hanging or cascading composition;
  soft flowing curves; common in Persian/Urdu calligraphy.

Base your answer only on visible visual style, not on the meaning of the text,
image background, decoration, or color.

If multiple styles seem possible, choose the closest match based on the
dominant letter shapes and composition.

Respond with only a single digit from 1 to 6. Do not explain.
\end{Verbatim}

\paragraph{Diwani-first prompt.}
\begin{Verbatim}[fontsize=\tiny,baselinestretch=0.95]
Look carefully at the calligraphic strokes, curves, and letterforms in this
Arabic calligraphy image. Based on the following characteristics, determine
which classical script family it belongs to:

Diwani: highly cursive and ornamental; dense flowing curves; letters may
interlace; often compact and decorative.
Kufic: angular, geometric, squared or block-like letterforms; strong
horizontal/vertical structure; often rigid and monumental.
Naskh: clear, balanced, readable manuscript style; rounded letters; moderate
stroke contrast; commonly used for Qur'anic and book text.
Thuluth: large, elegant, sweeping curves; elongated verticals; complex
decorative composition; often used in architectural or ceremonial calligraphy.
Ruq'ah: simple, fast handwriting-like style; compact letters; short strokes;
practical and less ornamental.
Nasta'liq: diagonal/slanted baseline; hanging or cascading composition; soft
flowing curves; common in Persian/Urdu calligraphy.

Base your answer only on visible visual style, not on the meaning of the text,
image background, decoration, or color. State only the name of the script."
\end{Verbatim}

\section{Representative confusion matrices}

\begin{figure}[H]
	\centering
	\includegraphics[width=0.92\linewidth]{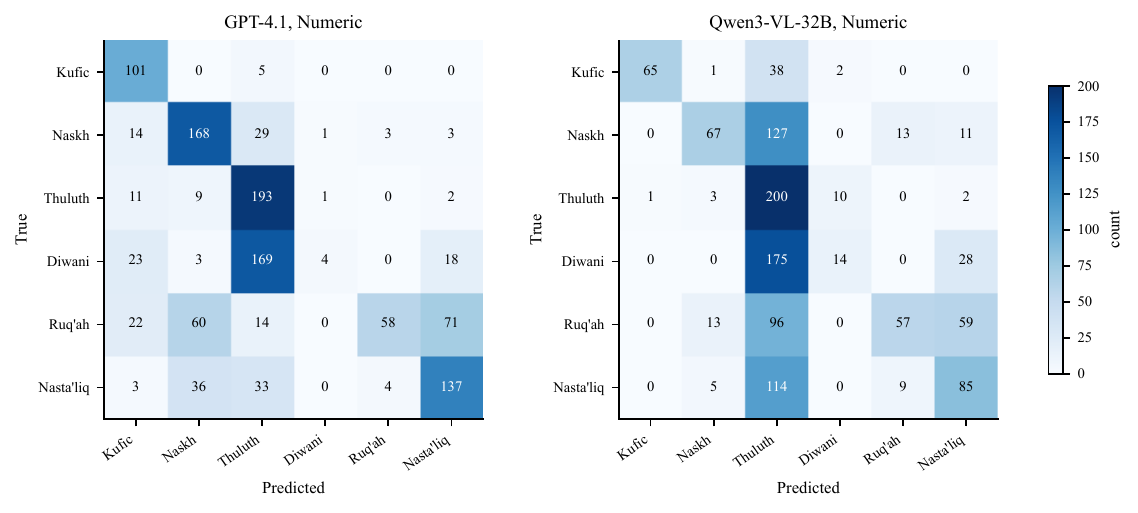}
	\caption{Representative numeric-prompt confusion matrices for GPT-4.1 and Qwen3-VL-32B.}
	\label{fig:representative_confusions}
\end{figure}

Figure~\ref{fig:representative_confusions} gives one closed/API and one open-weight numeric-prompt confusion matrix.
It complements the failure summary in Figure~\ref{fig:thuluth_diwani} by showing the full six-way redistribution of predictions.

\section{Per-style prompt profiles}

\begin{figure}[H]
\centering
\includegraphics[width=0.94\linewidth]{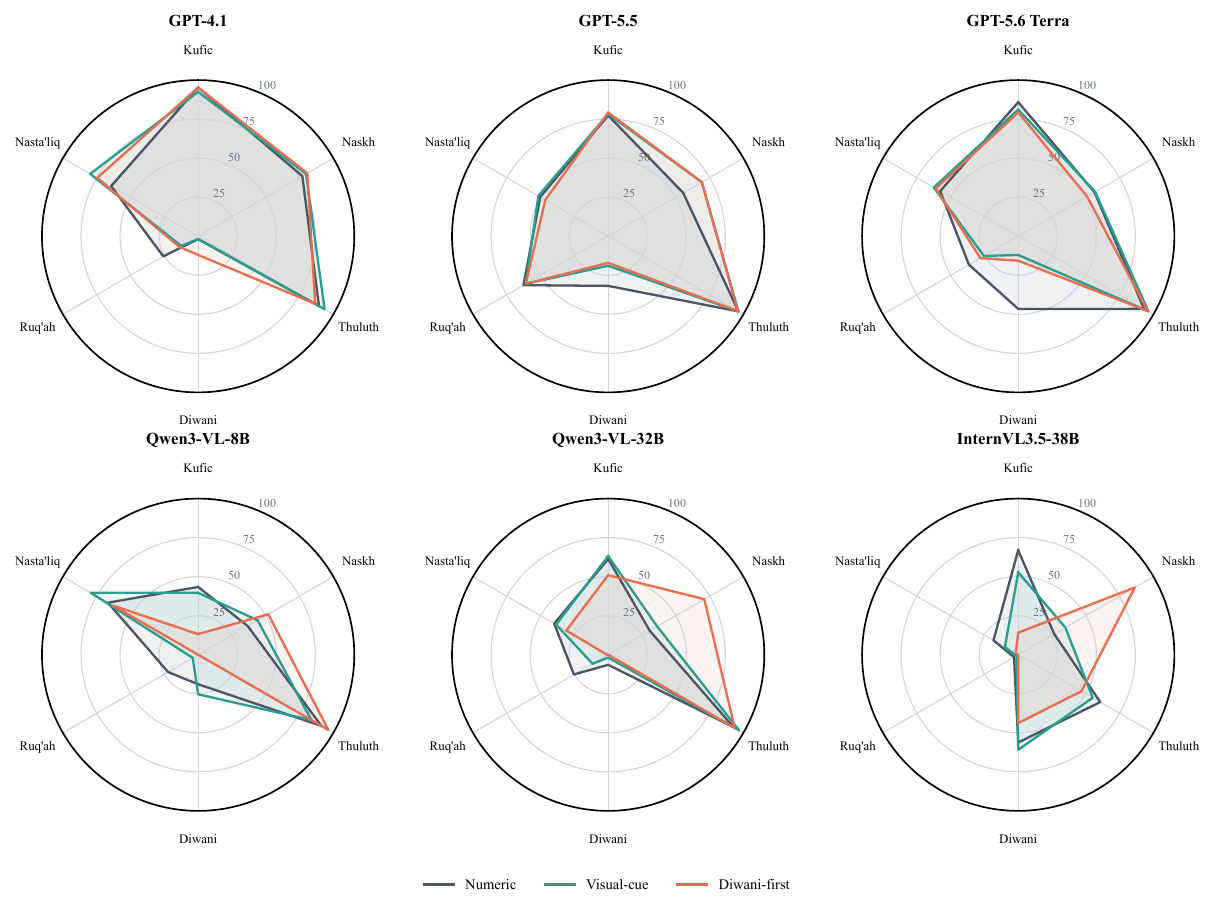}
\caption{Per-style recall across prompt conditions for six representative models.}
\label{fig:prompt_radars}
\end{figure}

Figure~\ref{fig:prompt_radars} shows per-style recall profiles for six representative models.
Each spoke is one calligraphy label, and each polygon is one prompt condition.
The profiles make prompt effects visible at the class level: some prompts recover Diwani for particular models, while others shift errors toward different labels rather than improving all styles uniformly.

\section{Full model--prompt summary}

Table~\ref{tab:full_runs} reports all 36 completed runs.
Values are generated from the retained closed/API CSV files and open-weight JSONL outputs.

\begin{table}[t]
\centering
\caption{Full model--prompt summary. Accuracy and balanced accuracy are percentages; Macro-F1 is unitless; Thuluth ratio is predicted count divided by true count; Diwani recall is a percentage. Bold marks best or closest-to-ideal metric values.}
\label{tab:full_runs}
\tiny
\setlength{\tabcolsep}{3.1pt}
\renewcommand{\arraystretch}{0.86}
\resizebox{\linewidth}{!}{
\rowcolors{2}{tablegray}{white}
\begin{tabular}{@{}llrrrrr@{}}
\toprule
Model & Prompt & Accuracy $\uparrow$ & Balanced acc. $\uparrow$ & Macro-F1 $\uparrow$ & Thuluth ratio & Diwani recall $\uparrow$ \\
\midrule
GPT-4o & Numeric & 50.5 & 54.1 & 0.504 & 2.93 & 15.7 \\
GPT-4o & Visual-cue & 60.8 & \textbf{63.3} & \textbf{0.627} & 2.02 & 43.3 \\
GPT-4o & Diwani-first & 59.6 & 62.6 & 0.607 & 1.56 & 41.0 \\
GPT-4.1 & Numeric & 55.3 & 58.9 & 0.507 & 2.05 & 1.8 \\
GPT-4.1 & Visual-cue & 56.6 & 60.0 & 0.516 & 2.11 & 1.8 \\
GPT-4.1 & Diwani-first & 56.9 & 60.5 & 0.527 & 1.86 & 12.0 \\
GPT-5.5 & Numeric & 60.9 & 62.3 & 0.616 & 2.05 & 31.8 \\
GPT-5.5 & Visual-cue & \textbf{61.1} & 62.5 & 0.615 & 2.19 & 18.9 \\
GPT-5.5 & Diwani-first & 59.8 & 61.5 & 0.603 & 2.24 & 17.1 \\
GPT-5.6 Terra & Numeric & 60.3 & 62.7 & 0.628 & 2.07 & 46.5 \\
GPT-5.6 Terra & Visual-cue & 53.1 & 55.7 & 0.533 & 2.55 & 12.0 \\
GPT-5.6 Terra & Diwani-first & 52.6 & 55.1 & 0.536 & 2.50 & 15.7 \\
Qwen2.5-VL-7B & Numeric & 29.2 & 29.1 & 0.255 & \textbf{0.91} & 9.2 \\
Qwen2.5-VL-7B & Visual-cue & 41.9 & 42.1 & 0.398 & 0.78 & 42.9 \\
Qwen2.5-VL-7B & Diwani-first & 47.6 & 46.9 & 0.438 & 0.54 & 84.8 \\
Qwen2.5-VL-32B & Numeric & 36.7 & 37.3 & 0.378 & 2.75 & 35.0 \\
Qwen2.5-VL-32B & Visual-cue & 43.3 & 43.1 & 0.410 & 2.60 & 14.3 \\
Qwen2.5-VL-32B & Diwani-first & 41.6 & 41.4 & 0.390 & 1.48 & 36.9 \\
Qwen3-VL-8B & Numeric & 46.5 & 46.5 & 0.457 & 2.69 & 18.9 \\
Qwen3-VL-8B & Visual-cue & 46.2 & 46.0 & 0.431 & 2.38 & 25.3 \\
Qwen3-VL-8B & Diwani-first & 39.2 & 37.4 & 0.292 & 2.89 & 0.0 \\
Qwen3-VL-32B & Numeric & 40.8 & 42.7 & 0.421 & 3.47 & 6.5 \\
Qwen3-VL-32B & Visual-cue & 39.1 & 41.4 & 0.385 & 3.65 & 1.8 \\
Qwen3-VL-32B & Diwani-first & 40.0 & 41.2 & 0.345 & 2.80 & 0.0 \\
InternVL3.5-8B & Numeric & 22.9 & 24.4 & 0.246 & 1.33 & 5.5 \\
InternVL3.5-8B & Visual-cue & 24.4 & 28.4 & 0.228 & 0.10 & 0.0 \\
InternVL3.5-8B & Diwani-first & 21.1 & 20.0 & 0.134 & 0.22 & \textbf{89.9} \\
InternVL3.5-38B & Numeric & 35.8 & 38.6 & 0.370 & 1.53 & 56.2 \\
InternVL3.5-38B & Visual-cue & 34.1 & 35.8 & 0.334 & 1.52 & 60.8 \\
InternVL3.5-38B & Diwani-first & 33.7 & 32.1 & 0.259 & \textbf{1.09} & 43.8 \\
MiniCPM-V 4.5 & Numeric & 31.6 & 30.8 & 0.331 & 1.17 & 42.4 \\
MiniCPM-V 4.5 & Visual-cue & 40.0 & 38.7 & 0.381 & 1.12 & 55.3 \\
MiniCPM-V 4.5 & Diwani-first & 20.2 & 19.6 & 0.111 & 4.94 & 0.5 \\
Gemma 4 12B-IT & Numeric & 47.2 & 49.4 & 0.454 & 2.89 & 6.5 \\
Gemma 4 12B-IT & Visual-cue & 44.1 & 46.1 & 0.419 & 3.28 & 0.5 \\
Gemma 4 12B-IT & Diwani-first & 43.8 & 45.4 & 0.415 & 3.35 & 0.0 \\
\bottomrule
\end{tabular}}
\end{table}

\section{Reproducibility notes}

All model--prompt runs use the same 1,195-row manifest.
Open-weight experiments use fixed seed 42 and greedy decoding.
Closed/API experiments were run with deterministic settings where supported; GPT-5-family models did not support the same temperature setting as GPT-4-family models, which limits exact reproducibility.
All closed/API CSVs and open-weight JSONL outputs are retained locally together with the script that regenerates summary metrics.

\section{Original DuwatBench-only context}

Before constructing the cleaned/rebalanced 1,195-image evaluation set, exploratory runs were conducted on the original DuwatBench~\safecitep{patle2026duwatbench}-style 1,272-row files.
Those files include 20 multi-label or ambiguous true-label rows; Table~\ref{tab:original_duwatbench} reports metrics after excluding them.
We do not mix these numbers into the main comparison because they use a different dataset distribution and fewer models.
They are nevertheless useful context: original DuwatBench is strongly Thuluth-heavy, so raw accuracy can look high even when balanced accuracy remains modest.

\begin{table}[t]
\centering
\caption{Original DuwatBench-only context runs on the clean six-class subset. These are not part of the main 36-run comparison. Bold marks the best value in each metric column.}
\label{tab:original_duwatbench}
\scriptsize
\setlength{\tabcolsep}{3.2pt}
\renewcommand{\arraystretch}{0.98}
\resizebox{\linewidth}{!}{
\rowcolors{2}{tablegray}{white}
\begin{tabular}{@{}llrrrrrr@{}}
\toprule
Model & Prompt/run & n & Accuracy $\uparrow$ & Balanced acc. $\uparrow$ & Macro-F1 $\uparrow$ & Thuluth ratio & Diwani recall $\uparrow$ \\
\midrule
GPT-4o & Closed-set B & 1252 & 65.9 & 45.9 & 0.475 & 1.27 & 30.8 \\
GPT-5.5 & Closed-set B & 1252 & 71.9 & 52.7 & 0.576 & 1.32 & 29.0 \\
GPT-4o & Alt wording C & 1252 & 49.3 & 41.6 & 0.368 & 0.55 & \textbf{85.3} \\
GPT-5.5 & Alt wording C & 1252 & 70.1 & 43.9 & 0.469 & 1.25 & 47.3 \\
GPT-4o & Label-order D & 1252 & 63.0 & 49.4 & 0.492 & \textbf{1.08} & 43.3 \\
GPT-5.5 & Label-order D & 1252 & \textbf{72.2} & \textbf{53.3} & \textbf{0.586} & 1.32 & 31.7 \\
GPT-5.5 & Pilot open-ended & 1252 & 69.5 & 50.1 & 0.538 & 1.09 & 59.4 \\
\bottomrule
\end{tabular}}
\end{table}

These context runs support two design choices in the main paper.
First, they motivate rebalancing: the original clean six-class subset has a 56.4\% majority-class baseline because Thuluth dominates the dataset.
Second, they show why prompt effects should be interpreted cautiously: GPT-4o's alternative wording run greatly increases Diwani recall while reducing overall accuracy, suggesting that prompt changes can swap the default label rather than solve fine-grained recognition.

\end{document}